# Optimizing Memory-Bounded Controllers for Decentralized POMDPs

Christopher Amato, Daniel S. Bernstein and Shlomo Zilberstein
Department of Computer Science
University of Massachusetts
Amherst, MA 01003
{camato,bern,shlomo}@cs.umass.edu

## Abstract

We present a memory-bounded optimization approach for solving infinite-horizon decentralized POMDPs. Policies for each agent are represented by stochastic finite state controllers. We formulate the problem of optimizing these policies as a nonlinear program, leveraging powerful existing nonlinear optimization techniques for solving the problem. While existing solvers only guarantee locally optimal solutions, we show that our formulation produces higher quality controllers than the state-of-the-art approach. We also incorporate a shared source of randomness in the form of a correlation device to further increase solution quality with only a limited increase in space and time. Our experimental results show that nonlinear optimization can be used to provide high quality, concise solutions to decentralized decision problems under uncertainty.

## 1 Introduction

Markov decision processes (MDPs) have been widely used to study single agent sequential decision making with full observability. Partially observable Markov decision processes (POMDPs) have had success modeling the more general situation in which the agent has only partial information about the state of the system. The decentralized partially observable Markov decision processes (DEC-POMDP) is an even more general framework which extends the POMDP model to mutiagent settings. In a DEC-POMDP each agent must make decisions based on uncertainty about the other agents as well as imperfect information of the system state. The agents seek to maximize a shared total reward using solely local information in order to act. Some examples of DEC-POMDP application areas include distributed robot control, networking and e-commerce.

Although there has been some recent work on exact and approximate algorithms for DEC-POMDPs (Nair et al., 2003; Bernstein et al., 2005; Hansen et al., 2004; Szer et al., 2005; Szer and Charpillet, 2005; Seuken and Zilberstein, 2007), only two algorithms (Bernstein et al., 2005; Szer and Charpillet, 2005) are able to find solutions for the infinite-horizon case. Such domains as networking and robot control problems, where the agents are in continuous use are more appropriately modeled as infinite-horizon problems. Exact algorithms require an intractable amount of space for all but the smallest problems. This may occur even if an optimal or near-optimal solution is concise. DEC-POMDP approximation algorithms can operate with a limited amount of memory, but as a consequence may provide poor results.

In this paper, we propose a new approach that addresses the space requirement of DEC-POMDP algorithms while maintaining a principled method based on the optimal solution. This approach formulates the optimal memory-bounded solution for the DEC-POMDP as a nonlinear program (NLP), thus allowing a wide range of powerful nonlinear optimization techniques to be applied. This is done by optimizing the parameters of fixed-size independent controllers for each agent, which when combined, produce the policy for the DEC-POMDP. While no existing NLP solver guarantees finding an optimal solution, our new formulation facilitates a more efficient search of the solution space and produces high quality controllers of a given size.

We also discuss the benefits of adding a shared source of randomness to increase the solution quality of our memory-bounded approach. This allows a set of independent controllers to be correlated in order to produce higher values, without sharing any local information. Correlation adds another mechanism in efforts to gain the most possible value with a fixed amount of space. This has been shown to be useful in order to increase value of fixed-size controllers (Bernstein *et al.*, 2005) and we show that is also useful when combined with our NLP approach.

The rest of the paper is organized as follows. We first present some background on the DEC-POMDP model and the finite state controller representation of their solution. We then describe the current infinite-horizon algorithms and describe some of their flaws. As an alternative, we present a nonlinear program that represents the optimal fixed-size solution. We also incorporate correlation into the NLP method and discuss its benefits. Lastly, experimental results are provided comparing the nonlinear optimization methods with and without correlation and the current state-of-the-art DEC-POMDP approximation algorithm. This is done by using an off-the-shelf, locally optimal nonlinear optimization method to solve the NLPs, but more sophisticated methods are also possible. For a range of domains and controller sizes, higher-valued controllers are found with the NLP and using correlation further increases solution quality. This suggests that high quality, concise controllers can be found in many diverse DEC-POMDP domains.

## 2　DEC-POMDP model and solutions

We first review the decentralized partially observable Markov decision process (DEC-POMDP) model. For clarity, we present the model for two agents as it is straightforward to extend it to $n$ agents.

A two agent DEC-POMDP can be defined with the tuple: $M = \langle S, A_1, A_2, P, R, \Omega_1, \Omega_2, O, T \rangle$

- $S$, the finite set of states
- $A_1$ and $A_2$, the finite sets of actions for each agent
- $P$, the set of state transition probabilities: $P(s'|s, a_1, a_2)$, the probability of transitioning from state $s$ to $s'$ when actions $a_1$ and $a_2$ are taken by agents 1 and 2 respectively
- $R$, the reward function: $R(s, a_1, a_2)$, the immediate reward for being in state $s$ and agent 1 taking action $a_1$ and agent 2 taking action $a_2$
- $\Omega_1$ and $\Omega_2$, the finite sets of observations for each agent
- $O$, the set of observation probabilities: $O(o_1, o_2|s', a_1, a_2)$, the probability of agents 1 and 2 seeing observations $o_1$ and $o_2$ respectively given agent 1 has taken action $a_1$ and agent 2 has taken action $a_2$ and this results in state $s'$

Since we are considering the infinite-horizon DEC-POMDP, the decision making process unfolds over an infinite sequence of stages. At each step, every agent chooses an action based on their local observation histories, resulting in an immediate reward and an observation for each agent. Note that because the state is not directly observed, it may be beneficial for the agent to remember the observation history. A *local policy* for an agent is a mapping from local observation histories to actions while a *joint policy* is a set of policies, one for each agent in the problem. The goal is to maximize the infinite-horizon total cumulative reward, beginning at some initial distribution over states called a *belief state*. In order to maintain a finite sum over the infinite-horizon, we employ a discount factor, $0 \le \gamma < 1$.

As a way to model DEC-POMDP policies with finite memory, finite state controllers provide an appealing solution. Each agent's policy can be represented as a local controller and the resulting set of controllers supply the joint policy, called the joint controller. Each finite state controller can formally be defined by the tuple $\langle Q, \psi, \eta \rangle$, where $Q$ is the finite set of controller nodes, $\psi : Q \to \Delta A$ is the action selection model for each node, and $\eta : Q \times A \times O \to \Delta Q$ represents the node transition model for each node given an action was taken and an observation seen. For $n$ agents, the value for starting in agent 1's nodes $\vec{q}$ and at state $s$ is given by

$$V(\vec{q}, s) = \sum_{\vec{a}} \prod_i^n P(a_i|q_i) \bigg[ R(s, \vec{a}) + \gamma \sum_{s'} P(s'|\vec{a}, s) \cdot \\ \sum_{\vec{o}} O(\vec{o}|s', \vec{a}) \sum_{\vec{q'}} \prod_i^n P(q_i'|q_i, a_i, o_i) V(\vec{q'}, s') \bigg]$$

This is also referred to as the Bellman equation. Note that the values can be calculated offline in order to determine controllers for each agent that can then be executed online for distributed control.

## 3　Previous work

As mentioned above, the only other algorithms that we know of that can solve infinite-horizon DEC-POMDPs are those of Bernstein et al. (2005) and Szer and Charpillet (2005). Bernstein et al.'s approach, called bounded policy iteration for decentralized POMDPs (DEC-BPI), is an approximate algorithm that also uses stochastic finite state controllers. Szer and Charpillet's approach is also an approximate algorithm, but it uses deterministic controllers.

DEC-BPI improves a set of fixed-size controllers by using linear programming. This is done by iterating through the nodes of each agent's controller and attempting to find an improvement. A linear program searches for a probability distribution over actions and transitions into the agent's current controller that increases the value of the controller for any initial state

For variables: $x(\vec{q}, \vec{a})$, $y(\vec{q}, \vec{a}, \vec{o}, \vec{q'})$ and $z(\vec{q}, s)$
Maximize
$$\sum_s b_0(s) z(\vec{q^0}, s)$$
Given the Bellman constraints:
$$\forall \vec{q}, s \ z(\vec{q}, s) = \sum_{\vec{a}} x(\vec{q}, \vec{a}) \left[ R(s, \vec{a}) + \gamma \sum_{s'} P(s'|s, \vec{a}) \sum_{\vec{o}} O(\vec{o}|s', \vec{a}) \sum_{\vec{q'}} y(\vec{q}, \vec{a}, \vec{o}, \vec{q'}) z(\vec{q'}, s') \right]$$

For each agent $i$ and set of agents, $-i$, apart from $i$,

Independence constraints:
$$\forall a_i, \vec{q} \ \sum_{a_{-i}} x(\vec{q}, \vec{a}) = \sum_{a_{-i}} x(q_i, q^f_{-i}, \vec{a})$$
$$\forall \vec{a}, \vec{q}, \vec{o}, q'_i \ \sum_{q'_{-i}} y(\vec{q}, \vec{a}, \vec{o}, \vec{q'}) = \sum_{q'_{-i}} y(q_i, q^f_{-i}, a_i, a^f_{-i}, o_i, o^f_{-i}, \vec{q'})$$

And probability constraints:
$$\forall q_i \ \sum_{\vec{a}} x(q_i, q^f_{-i}, \vec{a}) = 1 \text{ and } \forall q_i, o_i, a_i \ \sum_{\vec{q'}} y(q_i, q^f_{-i}, a_i, a^c_{-i}, o_i, o^f_{-i}, \vec{q'}) = 1$$
$$\forall \vec{q}, \vec{a} \ x(\vec{q}, \vec{a}) \geq 0 \text{ and } \forall \vec{q}, \vec{o}, \vec{a} \ y(\vec{q}, \vec{a}, \vec{o}, \vec{q'}) \geq 0$$

Table 1: The nonlinear program representing the optimal fixed-size controller. Variable $x(\vec{q}, \vec{a})$ represents $P(\vec{a}|\vec{q})$, variable $y(\vec{q}, \vec{a}, \vec{o}, \vec{q'})$ represents $P(\vec{q'}|\vec{q}, \vec{a}, \vec{o})$, variable $z(\vec{q}, s)$ represents $V(\vec{q}, s)$, $\vec{q^0}$ represents the initial controller node for each agent. Superscripted $f$'s such as $q^f_{-i}$ represent arbitrary fixed values.

and any initial node of the other agents' controllers. If the improvement is discovered, the node is updated based on the probability distributions found. Each node for each agent is examined in turn and the algorithm terminates when no agent can make any further improvements.

This algorithm allows memory to remain fixed, but provides only a locally optimal solution. This is due to the linear program considering the old controller values from the second step on and the fact that improvement must be over all possible states and initial nodes for the controllers of the other agents. As the number of agents or size of controllers grows, this later drawback is likely to severely hinder improvement.

Szer and Charpillet have developed a best-first search algorithm that finds deterministic finite state controllers of a fixed size. This is done by calculating a heuristic for the controller given the known deterministic parameters and filling in the remaining parameters one at a time in a best-first fashion. They prove that this technique will find the optimal deterministic finite state controller of a given size, but its use remains limited. This approach is very time and memory intensive and is restricted to deterministic controllers.

## 4 Nonlinear optimization approach

Due to the high space complexity of finding an optimal solution for a DEC-POMDP, fixed-size solutions are very appealing. Fixing memory balances optimality and computational concerns and should allow high quality solutions to be found for many problems. Using Bernstein et al.'s DEC-BPI method reduces problem complexity by fixing controller size, but solution quality is limited by a linear program that requires improvement across all states and initial nodes of the other agents. Also, each agent's controller is improved separately without consideration for the knowledge of the initial problem state, thus reducing solution quality. Both of these limitations can be eliminated by modeling a set of optimal controllers as a nonlinear program. By setting the value as a variable and using constraints to maintain validity, the parameters can be updated in order to represent the globally optimal solution over the infinite-horizon of the problem. Rather than the the iterative process of DEC-BPI, the NLP improves and evaluates the controllers of all agents at once for a given initial state in order to make the best possible use of the controller size.

Compared with other DEC-POMDP algorithms, the NLP approach makes more efficient use of memory than the exact methods, and using locally optimal NLP algorithms provides an approximation technique with a search based on the optimal solution of the problem. Rather than adding nodes and then attempting to remove those that will not improve the controller, as a dynamic programming approach might do, we search for the best controllers of a fixed size. The NLP is also able to take advantage of the start distribution, thus making better use of its size.

The NLP approach has already shown promise in the POMDP case. In a previous paper (Amato *et al.*, 2007), we have modeled the optimal fixed-size controller for a given POMDP as an NLP and with locally optimal solution techniques produced consistently higher quality controllers than a current state-of-the-art method. The success of the NLP in the single agent case suggested that an extension to DEC-POMDPs could also be successful. To construct this NLP, extra constraints are needed to guarantee independent controllers for each agent, while still maximizing the value.

### 4.1 Nonlinear problem model

The nonlinear program seeks to optimize the value of fixed-size controllers given a initial state distribution and the DEC-POMDP model. The parameters of this problem in vector notation are the joint action selection probabilities at each node of the controllers $P(\vec{a}|\vec{q})$, the joint node transition probabilities $P(\vec{q}'|\vec{q},\vec{a},\vec{o})$ and the values of each node in each state, $V(\vec{q},s)$. This approach differs from Bernstein et al.'s approach in that it explicitly represents the node values as variables. To ensure that the values are correct given the action and node transition probabilities, nonlinear constraints must be added to the optimization. These constraints are the Bellman equations given the policy determined by the action and transition probabilities. Constraints are also added to ensure distributed action selection and node transitions for each agent. We must also ensure that all probabilities are valid numbers between 0 and 1.

Table 1 describes the nonlinear program that defines the optimal controller for an arbitrary number of agents. The value of designated initial local nodes is maximized given the initial state distribution and the necessary constraints. The independence constraints guarantee that action selection and transition probabilities can be summed out for each agent by ensuring that they do not depend on any information that is not local.

**Theorem 1** *An optimal solution of the NLP results in optimal stochastic controllers for the given size and initial state distribution.*

**Proof sketch.** The optimality of the controllers follows from the NLP constraints and maximization of given initial nodes at the initial state distribution. The Bellman equation constraints restrict the value variables to valid amounts based on the chosen probabilities, the independence constraints guarantee distributed control and the maximum value is found for the initial nodes and state. Hence, this represents optimal controllers.

### 4.2 Nonlinear solution techniques

There are many efficient algorithms for solving large NLPs. When the problem is non-convex, as in our case, there are multiple local maxima and no NLP solver guarantees finding the optimal solution. Nevertheless, existing techniques proved useful in finding high-quality results for large problems.

For this paper, we used a freely available nonlinearly constrained optimization solver called *filter* (Fletcher *et al.*, 2002) on the NEOS server (http://www-neos.mcs.anl.gov/neos/). Filter finds solutions by a method of successive approximations called sequential quadratic programming (SQP). SQP uses quadratic approximations which are then more efficiently solved with quadratic programming (QP) until a solution to the more general problem is found. A QP is typically easier to solve, but must have a quadratic objective function and linear constraints. Filter adds a "filter" which tests the current objective and constraint violations against those of previous steps in order to promote convergence and avoid certain locally optimal solutions. The DEC-POMDP and nonlinear optimization models were described using a standard optimization language AMPL.

## 5 Incorporating correlation

Bernstein et al. also allow each agent's controller to be correlated by using a shared source of randomness in the form of a *correlation device*. As an example of one such device, imagine that before each action is taken, a coin is flipped and both agents have access to the outcome. Each agent can then use that new information to affect their choice of action. Along with stochasticity, correlation is another means of increasing value when memory is limited.

A correlation device provides extra signals to the agents and operates independently of the DEC-POMDP. That is, the correlation device is a tuple $\langle C, \psi \rangle$, where $C$ is a set of states and $\psi : C \to \Delta C$ is a stochastic transition function that we will represent as $P(c'|c)$. At each step of the problem, the device transitions and each agent can observe its state.

For variables: $w(c,c'), x(\vec{q},\vec{a},c), y(\vec{q},\vec{a},\vec{o},\vec{q'},c)$ and $z(\vec{q},s,c)$
Maximize
$$\sum_s b_0(s) z(\vec{q^0}, s)$$
Given the Bellman constraints:
$$\forall \vec{q}, s \; z(\vec{q},s,c) = \sum_{\vec{a}} x(\vec{q},\vec{a},c) \left[ R(s,\vec{a}) + \gamma \sum_{s'} P(s'|s,\vec{a}) \sum_{\vec{o}} O(\vec{o}|s',\vec{a}) \sum_{\vec{q'}} y(\vec{q},\vec{a},\vec{o},\vec{q'},c) \sum_{c'} w(c,c') z(\vec{q'},s',c') \right]$$

Table 2: The nonlinear program representing the optimal fixed-size controller including a correlation device. Variable $x(\vec{q},\vec{a},c)$ represents $P(\vec{a}|\vec{q},c)$, variable $y(\vec{q},\vec{a},\vec{o},\vec{q'},c)$ represents $P(\vec{q'}|\vec{q},\vec{a},\vec{o},c)$, variable $z(\vec{q},s,c)$ represents $V(\vec{q},s,c)$, $\vec{q^0}$ represents the initial controller node for each agent and $w(c,c')$ represents $P(c'|c)$. The other constraints are similar to those above with the addition of a sum to one constraint for the correlation device.

The independent local controllers defined above can be modified to make use of the correlation device. This is done by making the parameters dependent on the signal from the correlation device. For agent $i$, action selection is then $P(a_i|q_i,c)$ and node transition is $P(q_i'|q_i,a_i,o_i,c)$. For $n$ agents, the value of the correlated joint controller beginning in nodes $\vec{q}$, state $s$ and correlation device state $c$ is defined as $V(\vec{q},s,c) =$

$$\sum_{\vec{a}} \prod_i^n P(a_i|q_i,c) \left[ R(s,\vec{a}) + \gamma \sum_{s'} P(s'|\vec{a},s) \sum_{\vec{o}} O(\vec{o}|s',\vec{a}) \cdot \sum_{\vec{q'}} \prod_i^n P(q_i|q_i,a_i,o_i,c) \sum_{c'} P(c'|c) V(\vec{q'},s',c') \right]$$

Our NLP can be extended to include a correlation device. This optimization problem, the first part of which is shown in Table 2, is very similar to the previous NLP. A new variable is added for the transition function of the correlation device and the other variables now include the signal from the device. The Bellman equation incorporates the new correlation device signal at each step, but the other constraints remain the same. A new probability constraint is also added to ensure that the transition probabilities for each state of the correlation device sum to one.

## 6 Experimental results

We tested our nonlinear programming approach in three DEC-POMDP domains. In each experiment, we compare Bernstein et al.'s DEC-BPI with NLP solutions using filter for a range of controller sizes. We also implemented each of these approaches with a correlation device of size two. We do not compare with Szer and Charpillet's algorithm because the problems presented in that work are slightly different than those used by Bernstein et al. Nevertheless, on the problems that we tested, our approach can and does achieve higher values than Szer and Charpillet's algorithm for all of the controller sizes for which that the best-first search is able to find a solution.

| size | DEC-BPI | DEC-BPI corr | NLO | NLO-corr |
|------|---------|--------------|-----|----------|
| 1    | 4.687   | 6.290        | 9.1 | 9.1      |
| 2    | 4.068   | 7.749        | 9.1 | 9.1      |
| 3    | 8.637   | 7.781        | 9.1 | 9.1      |
| 4    | 7.857   | 8.165        | 9.1 | 9.1      |

Table 3: Broadcast problem values using NLP methods and DEC-BPI with and without a 2 node correlation device

| size | DEC-BPI | DEC-BPI corr | NLO   | NLO-corr |
|------|---------|--------------|-------|----------|
| 1    | < 1s    | < 1s         | 1s    | 2s       |
| 2    | < 1s    | 2s           | 3s    | 8s       |
| 3    | 2s      | 7s           | 764s  | 2119s    |
| 4    | 5s      | 24s          | 4061s | 10149s   |

Table 4: Broadcast problem mean optimization times using NLP methods and DEC-BPI with and without a 2 node correlation device

Each NLP and DEC-BPI algorithm was run until convergence was achieved with ten different random deterministic initial controllers, and the mean values and times are reported. The times reported for each NLP method can only be considered estimates due to running each algorithm on external machines with uncontrollable load levels, but we expect that they vary by only a small constant. Note that our goal in these experiments is to demonstrate the benefits of our formulation when used in conjunction with an "off the shelf" solver such as filter. The formulation is very general and many other solvers may be applied. Throughout this section we will refer to our nonlinear optimization as NLO and the optimization with the correlation device with two states as NLO-corr.

### 6.1 Broadcast problem

A DEC-POMDP used by Bernstein et al. was a simplified two agent networking example. This problem has 4 states, 2 actions and 5 observations. At each time step, each agent must choose whether or not to send a message. If both agents send, there is a collision and neither gets through. A reward of 1 is given for every step a message is successfully sent over the channel and all other actions receive no reward. Agent 1 has a 0.9 probability of having a message in its queue on each step and agent 2 has only a 0.1 probability. The domain is initialized with only agent 1 possessing a message and a discount factor of 0.9 was used.

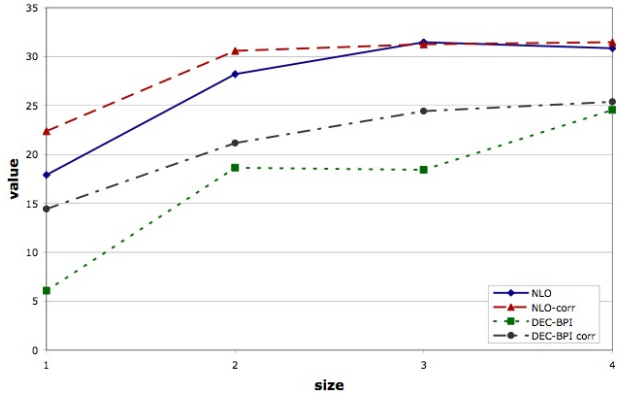

Figure 1: Recycling robots values using NLP methods and DEC-BPI with and without a 2 node correlation device

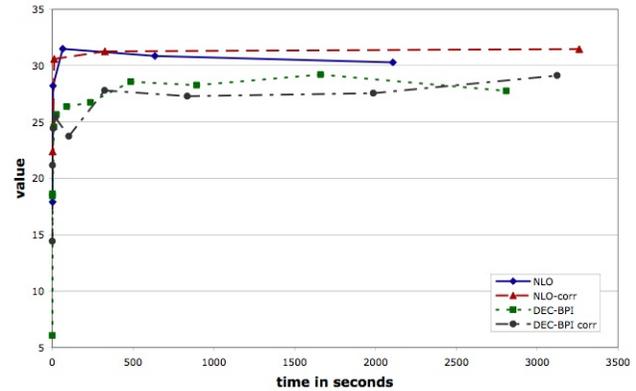

Figure 2: Recycling robots graphs for value vs time for the NLP and DEC-BPI methods with and without the correlation device.

Table 3 shows the values produced by DEC-BPI and our nonlinear programming approach with and without a correlation device for several controller sizes. Both nonlinear techniques produce the same value, 9.1 for each controller size. In all cases this is a higher value than that produced by Bernstein et al.'s independent and correlated approaches. As 9.1 is the maximum value that any approach that we tested receives for the given controller sizes, it is likely that it is optimal for these sizes.

The time used by each algorithm is shown in Table 4. As expected, the nonlinear optimization methods require more time to find a solution than the DEC-BPI methods. As noted above, solution quality is also higher using nonlinear optimization. Either NLP approach can produce a higher valued one node controller in an amount of time similar to or less than each DEC-BPI method. Therefore, for this problem, the NLP methods are able to find higher valued, more concise solutions given a fixed amount of space or time.

### 6.2 Recycling robots

As another comparison, we have extended the Recycling Robot problem (Sutton and Barto, 1998) to the multiagent case. The robots have the task of picking up cans in an office building. They have sensors to find a can and motors to move around the office in order to look for cans. The robots are able to control a gripper arm to grasp each can and then place it in an on-board receptacle. Each robot has three high level actions: (1) search for a small can, (2) search for a large can or (3) recharge the battery. In our two agent version, the larger can is only retrievable if both robots pick it up at the same time. Each agent can decide to independently search for a small can or to attempt to cooperate in order to receive a larger reward. If only one agent chooses to retreive the large can, no reward is given. For each agent that picks up a small can, a reward 2 is given and if both agents cooperate to pick the large can, a reward of 5 is given. The robots have the same battery states of high and low, with an increased likelihood of transitioning to a low state or exhausting the battery after attempting to pick up the large can. Each robot's battery power depends only on its own actions and each agent can fully observe its own level, but not that of the other agent. If the robot exhausts the battery, it is picked up and plugged into the charger and then continues to act on the next step with a high battery level. The two robot version used in this paper has 4 states, 3 actions and 2 observations. A discount factor of 0.9 was used.

We can see in Figure 1 that in this domain higher quality controllers are produced by using nonlinear optimization. Both NLP methods permit higher mean values than either DEC-BPI approach for all controller sizes. Also, correlation is helpful for both the NLP and DEC-BPI approaches, but becomes less so for larger controller sizes. For the nonlinear optimization cases, both approaches converge to within a small amount of the maximum value that was found for any controller size tested. As controller size grows, the NLP methods are able to reliably find this solution and correlation is no longer useful.

The running times of each algorithm follow the same trend as above in which the nonlinear optimization approaches required much more time as controller size increases. The ability for the NLP techniques to produce smaller, higher valued controllers with similar or lesser running time also follows the same trend.

Figure 2 shows the values that can be attained for each method based on the mean time necessary for convergence. Results are included for NLP techniques up to four nodes with the correlation device and five nodes without it while DEC-BPI values are given for fourteen nodes with the correlation device and eighteen without it. This graph demonstrates that even if we allow controller size to continue to grow and examine only the amount of time that is necessary to achieve a solution, the NLP methods continue to provide higher values. Although the values of the controllers produced by the DEC-BPI methods are somewhat close to those of the NLP techniques as controller size grows, our approaches produce that value with a fraction of the controller size.

### 6.3 Multiagent tiger problem

Another domain with 2 states, 3 actions and 2 observations called the multiagent tiger problem was introduced by Nair et al. (Nair *et al.*, 2003). In this problem, there are two doors. Behind one door is a tiger and behind the other is a large treasure. Each agent may open one of the doors or listen. If either agent opens the door with the tiger behind it, a large penalty is given. If the door with the treasure behind it is opened and the tiger door is not, a reward is given. If both agents choose the same action (i.e., both opening the same door) a larger positive reward or a smaller penalty is given to reward this cooperation. If an agent listens, a small penalty is given and an observation is seen that is a noisy indication of which door the tiger is behind. While listening does not change the location of the tiger, opening a door causes the tiger to be placed behind one of the door with equal probability. A discount factor of 0.9 was used.

Figure 3 shows the values attained by each NLP and DEC-BPI method for the given controller sizes. Figure 4 shows the values of just the two NLP methods. These graphs show that not only do the NLP methods significantly outperform the DEC-BPI approaches, but correlation greatly increases the value attained by the nonlinear optimization. The individual results for this problem suggest the DEC-BPI approach is more dependent on the initial controller and the large penalties in this problem result in several results that are very low. This outweighs the few times that more reasonable value is attained. Nevertheless, the max value attained by DEC-BPI for all cases is still less than the mean value attained by the NLP methods. Again for this problem, more time is needed for the NLP approaches, but one node controllers are produced with higher value than any controller size for the DEC-BPI methods and require very little time.

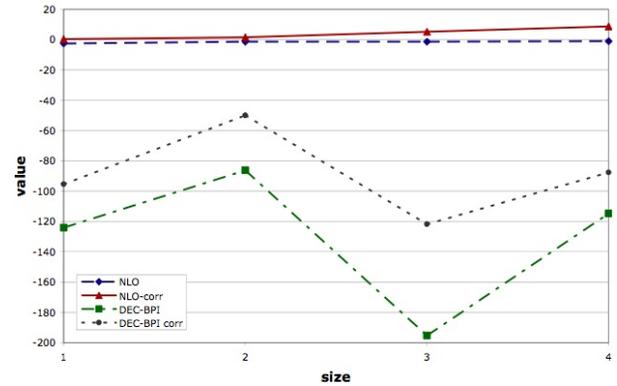

Figure 3: Multiagent Tiger problem values using NLP methods and DEC-BPI with and without a 2 node correlation device.

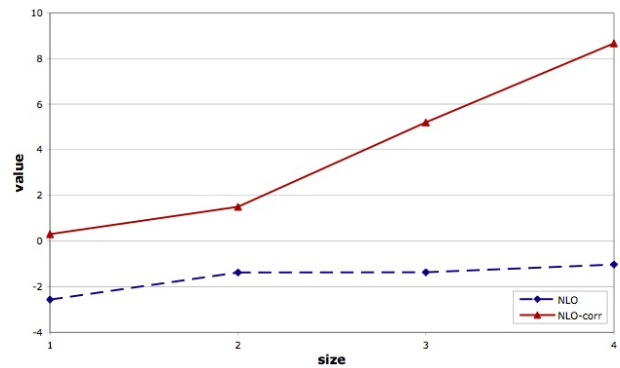

Figure 4: Multiagent Tiger problem values using just the NLP methods with and without a 2 node correlation device.

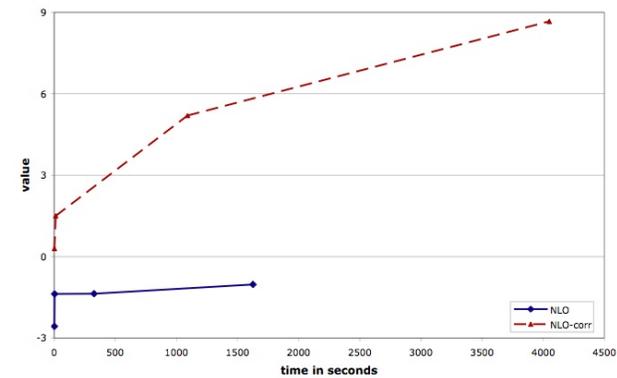

Figure 5: Multiagent Tiger problem graphs for value vs. time for the NLP methods with and without the correlation device.

The usefulness of the correlation device is illustrated in Figure 5. For given amounts of time, the nonlinear optimization that includes the correlation device produces much higher values. The DEC-BPI methods are not included in this graph as they were unable to produce mean values greater than -50 for any controller size up to 22 for which mean time to convergence was over 5000 seconds. This shows the importance of correlation in this problem and the ability of our NLP technique to take advantage of it.

## 7 Conclusion

We introduced a novel approach to solving decentralized POMDPs by using a nonlinear program formulation. This memory-bounded stochastic controller formulation allows a wide range of powerful nonlinear programming algorithms to be applied to solve DEC-POMDPs. The approach is easy to implement as it mostly involves reformulating the problem and feeding it into an NLP solver.

We showed that by using an off-the-shelf locally optimal NLP solver, we were able to produce higher valued controllers than the current state-of-the-art technique for an assortment of DEC-POMDP problems. Our experiments also demonstrate that incorporating a correlation device as a shared source of randomness for the agents can further increase solution quality. While the time taken to find a solution to the NLP can be higher, the fact that higher values can be found with smaller controllers by using the NLP suggests adopting more powerful optimization techniques for smaller controllers can be more productive in a given amount of time. The combination of start state knowledge and more advanced optimization allows us to make efficient use of the limited space of the controllers. These results show that this method can allow compact optimal or near-optimal controllers to be found for various DEC-POMDPs.

In the future, we plan to conduct a more exhaustive analysis of the NLP representation and explore more specialized algorithms that can be tailored for this optimization problem. While the performance we get using a standard nonlinear optimization algorithm is very good, specialized solvers might be able to further increase solution quality and scalability. We also plan to characterize the circumstances under which introducing a correlation device is cost effective.

**Acknowledgements**

An earlier version of this paper without improvements such as incorporating a correlation device was presented at the AAMAS-06 Workshop on Multi-Agent Sequential Decision Making in Uncertain Domains. This work was supported in part by the Air Force Office of Scientific Research (Grant No. FA9550-05-1-0254) and by the National Science Foundation (Grant No. 0535061). Any opinions, findings, conclusions or recommendations expressed in this manuscript are those of the authors and do not reflect the views of the US government.